\documentclass{youtu}

\usepackage{mathpazo}
\usepackage{amsmath,amssymb,amsfonts,mathtools}
\usepackage[T1]{fontenc}
\usepackage[authoryear,round]{natbib}
\usepackage{wrapfig}
\usepackage{colortbl}
\definecolor{dspinerow}{HTML}{FBEEE3}  %
\usepackage[capitalize,noabbrev]{cleveref}
\DeclareMathOperator{\softmax}{softmax}

\newcommand{\cL}{\mathcal{L}}

\DeclareMathOperator{\RMS}{RMS}
\DeclareMathOperator{\normalize}{normalize}
\DeclareMathOperator{\sg}{sg}

\title{Draft in Parallel, Condition Through Depth: Adjacent Causal Injection for Speculative Decoding}
\author{Haohui~Zhang$^{1,2,\dagger}$, Keyu~Chen$^{2,\dagger,\ddagger}$, Haocheng~Sun$^{3}$, Weibo~Gu$^{2}$, Ruizhi~Qiao$^{2}$, Xing~Sun$^{2}$, Bo~Jiang$^{1,*}$}
\affiliation{$^{1}$Shanghai Jiao Tong University\qquad $^{2}$Tencent YouTu Lab\qquad $^{3}$Xiamen University}
\renewcommand{\contributionlist}{\begin{center}\small $^{\dagger}$Equal contribution.\qquad $^{\ddagger}$Project lead.\qquad $^{*}$Corresponding author.\end{center}}
\hypersetup{
  pdftitle={Draft in Parallel, Condition Through Depth: Adjacent Causal Injection for Speculative Decoding},
  pdfauthor={Haohui Zhang, Keyu Chen, Haocheng Sun, Weibo Gu, Ruizhi Qiao, Xing Sun, Bo Jiang},
  pdfsubject={Parallel speculative decoding},
  pdfkeywords={DSpine, speculative decoding, causal conditioning}
}

\abstract{
Parallel speculative drafting generates multiple candidates in one backbone pass, but independent token selection can produce inconsistent continuations that shorten the accepted prefix. Existing methods mostly leave conditional decoding to a lightweight module after the backbone, which limits the flow of predecessor information to successors. Our analysis of DFlash shows that early positions already form recoverable predictions in shallow layers, and that accurate adjacent predecessors help successors more when they enter earlier. We therefore propose DSpine, a drafter with causal conditioning injection throughout the backbone: at every layer, gated adjacent injection writes each predecessor's predicted feature into its successor, so the causal conditioning chain unfolds over network depth while all positions update in parallel. A unified transfer space built from the target model's output embeddings unifies layer-wise injection with predecessor-conditioned decoding, and layer-wise output-embedding supervision promotes the formation of predicted features in shallow layers. Fused kernels and a transition cache execute both efficiently in parallel within SGLang. Across seven math, code, and chat benchmarks, DSpine achieves the longest acceptance length at both temperatures on Qwen3-4B and Qwen3-8B. At temperature zero on Qwen3-8B, it raises the seven-benchmark mean from DFlash's 3.77 to 4.82 (+27.8\%); in SGLang serving tests, it delivers 23.3\% higher throughput than DFlash on average.
}
\providecommand{\project}[1]{}
\project{https://dspine-project.github.io}

\begin{document}

\maketitle

\par\vspace{5pt}
\noindent\begin{minipage}{\linewidth}
\centering
\includegraphics[width=0.96\linewidth]{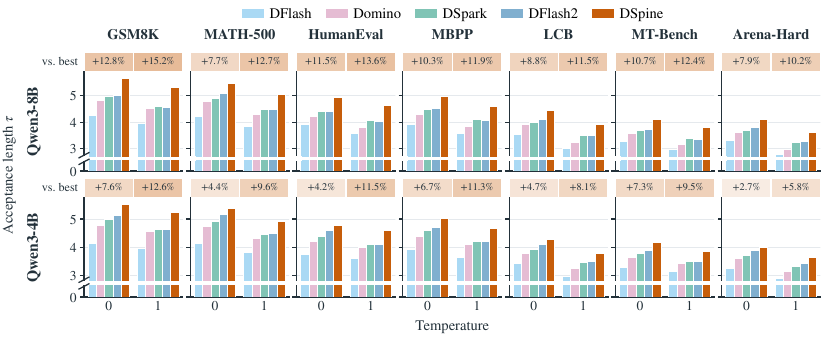}
\captionof{figure}{Mean acceptance length across seven math, code, and chat benchmarks. DSpine leads all compared parallel drafters on Qwen3-8B and Qwen3-4B at temperatures 0 and 1; the annotations report its margin over the strongest baseline in each setting.}
\label{fig:headline-acceptance}
\end{minipage}
\par\vspace{5pt}

\section{Introduction}
\label{sec:intro}

Autoregressive decoding is a major source of latency in large language models. Speculative decoding accelerates it without changing the target model: a lightweight drafter proposes multiple candidates, and the target model verifies them in a single forward pass \citep{leviathan2023fast}. Its speedup depends on how many candidates are accepted per round and how cheaply they are drafted. Autoregressive drafters such as EAGLE-3 reuse target-model features to improve predictions but still require one sequential drafting step per draft position \citep{li2025eagle3}. Inspired by diffusion language modeling, DFlash instead predicts a masked block in parallel within one forward pass conditioned on target context features, so a longer draft adds no sequential drafting steps \citep{chen2026dflash}.

This shift also changes how token dependencies are established. Autoregressive drafting conditions each successor on its already selected predecessors, whereas positions in a parallel draft must form predictions before their predecessors commit to tokens, so individually plausible candidates can be inconsistent when combined. Because the target accepts candidates in prefix order, a single inconsistency truncates the rest of the round's draft, leaving much of the cheaply drafted block unaccepted. Reducing drafting cost must therefore go hand in hand with establishing effective predecessor conditioning inside parallel computation.

Recent work improves token consistency in parallel drafts through lightweight conditioning modules. Inside the draft backbone, DFlash2 mixes adjacent positions through local convolutions, but these blend neighboring hidden states without separating the predecessor's predictive information from the rest of its state \citep{inco2026dflash2}. Domino, DSpark, and DFlash2's path selector instead condition on the predecessor after the backbone to adjust token distributions or candidate selection \citep{huang2026domino,cheng2026dspark,inco2026dflash2}. Conditional decoding in these methods is thus left to a lightweight module after the backbone, which limits the flow of predecessor information to successors. Moreover, the predecessor information these modules pass is confined to the final-layer vocabulary space, so it can neither be combined with in-layer features nor exploit the rich intermediate features formed across the backbone.

Our empirical study yields two key insights. \textbf{(1) Earlier positions form predictions at shallower depth:} layer-wise readouts already recover substantial token-predictive information at early positions in shallow layers (\cref{fig:conditioning_observations}(a)), so the prefix could condition successors whose computation is still in progress. \textbf{(2) Successor prediction depends on both the content of the adjacent predecessor and when it enters the network:} even with correct earlier context, replacing only the adjacent predecessor with its draft prediction sharply lowers successor accuracy, and injecting the same correct predecessor features earlier clearly improves it (\cref{fig:conditioning_observations}(c)).

Guided by these insights, we propose \textbf{DSpine}, a drafter with causal conditioning injection throughout the backbone. At every layer, a gated adjacent injection module writes each predecessor's predicted feature into its successor, so that the causal conditioning chain unfolds over network depth while all positions update in parallel. We further build a unified transfer space from the output embeddings of the target model, which unifies layer-wise injection with the information transfer in predecessor-conditioned decoding. We also design layer-wise output-embedding supervision that aligns the predicted features with this space, promoting their formation in shallow layers and reducing their discrepancy from token embeddings. Finally, we introduce a transition cache that precomputes the scores of all adjacent candidate pairs in parallel, turning sequential conditional decoding into parallel computation. Our contributions are:
\begin{itemize}
    \item We identify recoverable shallow prefix predictions and the effect of predecessor-conditioning timing, motivating dependency formation across network depth.
    \item We propose DSpine, which injects causal conditioning throughout the backbone and links it to predecessor-conditioned decoding through a unified output-embedding space with layer-wise supervision.
    \item We demonstrate that DSpine surpasses baselines trained on the same data in both acceptance length and SGLang throughput: on Qwen3-8B, it raises the mean acceptance length from 3.77 for DFlash to 4.82 (\cref{fig:headline-acceptance}), and its SGLang throughput exceeds that of DFlash by 23.3\%.
\end{itemize}

\section{Preliminaries}
\label{sec:prelim}

\subsection{Speculative Decoding}
\label{sec:prelim-spec}

Given a confirmed context $x_{\le 0}$, a drafter proposes $M$ candidates $\hat{x}_{1:M}$, which the target model verifies in one forward pass \citep{leviathan2023fast}. Let $q_{\mathrm{draft},t}$ be the actual proposal distribution given this context and preceding candidates; both it and the target distribution $p_t$ include their respective sampling transformations. Conditional on acceptance of all preceding candidates, candidate $t$ is accepted with probability
\begin{equation}
\min\!\left(1,\frac{p_t(\hat{x}_t)}{q_{\mathrm{draft},t}(\hat{x}_t)}\right),
\qquad
p_t(v)=p_{\mathrm{target}}(v\mid x_{\le 0},\hat{x}_{1:t-1}).
\label{eq:spec-accept}
\end{equation}
At the first rejection, the remaining draft is discarded and a replacement is sampled from the corrected distribution (\cref{eq:spec-correction}); if all candidates pass, the target supplies an additional token. Greedy decoding accepts candidates matching the target's choice under the same prefix.

The average acceptance length $\tau=\mathbb E[N_{\mathrm{acc}}]+1$ is the number of tokens advanced per round, where $N_{\mathrm{acc}}$ counts consecutively accepted draft tokens and the extra one is the correction or additional target token. For a fixed runtime configuration, ignoring prefill and termination costs, the average time per output token (TPOT) is approximately
\begin{equation}
\mathrm{TPOT}_{\mathrm{spec}}\approx
\frac{T_{\mathrm{draft}}+T_{\mathrm{verify}}}{\tau},
\label{eq:spec-speedup}
\end{equation}
where $T_{\mathrm{draft}}$ and $T_{\mathrm{verify}}$ are mean per-round costs, including sampling and cache management. Later candidates contribute only when the preceding prefix is accepted, and acceptance gains must compensate for added drafting cost.

\subsection{Parallel Drafting}
\label{sec:prelim-parallel}

DFlash-style parallel drafters process a block of $M+1$ positions, the known anchor $x_0$ followed by $M$ masked candidate positions \citep{chen2026dflash,zhang2026dflare}. In DFlash, fused target-model features over $x_{<0}$ supply K/V at every layer. The backbone updates all positions in one pass with bidirectional block attention, and the LM head shared with the target produces candidate distributions.

An autoregressive drafter generates each token conditioned on the predecessors it has drafted, whereas a DFlash-style drafter predicts every position from the confirmed context alone, so the block proposal factorizes over positions:
\begin{equation}
q_{\mathrm{draft}}(\hat{x}_{1:M}\mid x_{\le 0})
=\prod_{t=1}^{M}q_{\mathrm{draft},t}(\hat{x}_t\mid x_{\le 0}).
\label{eq:parallel-proposal}
\end{equation}
Each position thus marginalizes over its possible predecessors instead of conditioning on the one actually drafted, and individually likely tokens can form inconsistent combinations. Because verification accepts candidates in prefix order, such inconsistencies cut the accepted prefix, and acceptance decays rapidly at later positions \citep{cheng2026dspark}.

\section{When and How Prefix Predictions Condition Parallel Drafting}
\label{sec:obs}

This section asks whether evolving prefix predictions can condition successor computation in parallel drafting. We first examine when predictive information becomes recoverable and when reliable prefix conditions are most useful, then turn to the immediate predecessor: its contribution beyond earlier history and the timing of its use. We use the frozen official Qwen3-8B-DFlash-b16 with greedy Qwen3-8B continuations as the correct tokens; datasets, metrics, and full results are in \cref{app:obs}.

\begin{figure}[t]
\centering
\includegraphics[width=\linewidth]{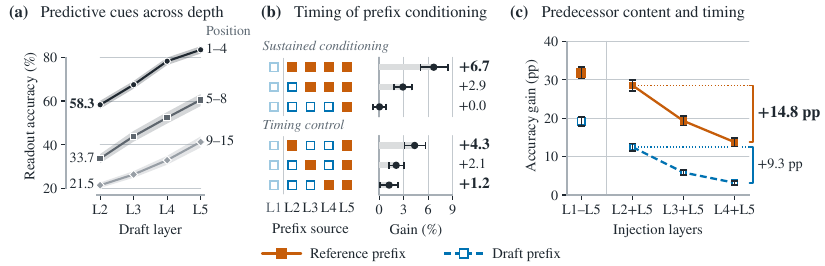}
\caption{(a) Readout accuracy of each layer, grouped by block position. (b) Gain in consecutive correct tokens when the drafted prefix is replaced by the correct prefix at the filled (orange) layers; top: from one layer onward; bottom: at two layers, moving the earlier one. (c) Gain in successor accuracy when, at the specified layers, the adjacent predecessor's features come from the correct token rather than its draft prediction (L1--L5: all layers); line color marks whether the history before the predecessor is correct (orange) or drafted (blue).}
\label{fig:conditioning_observations}
\label{fig:prefix_content_timing}
\label{fig:predecessor_content_timing}
\end{figure}

\subsection{Early Predictive Cues and the Timing of Conditioning}
\label{sec:obs-shallow}

To examine when predictive information emerges, we read out predictions from each layer. Intermediate DFlash states lie outside the input space of the target LM head and cannot be decoded directly, so for each layer we fit, on separate data, an equal-capacity lightweight low-rank projection that maps its frozen states into the space decoded by the frozen LM head. At the same shallow depth, early positions are read out much more accurately than distant ones, while all position groups improve with depth (\cref{fig:prefix_content_timing}(a)). Although positions update in parallel, predictive information is not equally available: prefixes already carry recoverable cues while successor predictions are still forming, so prefixes could condition successor computation before selecting their own tokens.

We next examine when more accurate prefix conditions are most useful. With a correct token as the anchor, we fill the preceding prefix with draft predictions or correct tokens, which the target model encodes for DFlash's existing context pathway. Sustained access to correct-prefix features from an early layer yields a 6.7\% relative gain in consecutive correct tokens over the all-draft-prefix baseline, and delaying it reduces the gain (\cref{fig:prefix_content_timing}(b), sustained conditioning).

Earlier access, however, also means that more layers receive correct features. To isolate timing, we fix the total number of injection layers and vary only their placement: earlier injection raises the relative gain from 1.2\% to 4.3\% (\cref{fig:prefix_content_timing}(b), timing control). Thus, even at a fixed number of uses, earlier access to reliable conditions improves successor prediction: a reliable prefix is valuable not only for more accurate content but also for timely use as successor representations form.

\subsection{Adjacent Predecessors Shape Successor Predictions}
\label{sec:obs-adjacent}

We next examine how the immediate predecessor affects successor prediction. Keeping the target-encoded earlier history correct, we use the predecessor as the anchor of a new draft block: replacing only the correct predecessor with its draft prediction lowers successor accuracy by 31.8 percentage points on average over three block positions (L1--L5 in \cref{fig:predecessor_content_timing}(c)). Correct earlier history thus does not fully compensate for an inaccurate predecessor: the adjacent token supplies additional predictive cues for its successor.

The predecessor's effect also depends on when it enters successor computation. Keeping the draft-predicted anchor, we replace its K/V at selected layers with the same-layer K/V cached from a forward pass with the correct predecessor. With correct earlier history, the same number of injection layers, and the same predecessor K/V at the final layer, earlier injection improves successor accuracy over later injection by 14.8 points on average (\cref{fig:predecessor_content_timing}(c)), indicating that adjacent conditions affect how successor representations form.

These two observations are complementary: shallow prefix states already carry recoverable predictive cues, and accurate predecessor features help successors more when supplied earlier. Motivated by this, we explore transmitting predecessor prediction features to successors from shallow layers, letting token dependencies build up progressively through network depth (\cref{sec:method}).

\section{DSpine: Adjacent Causal Injection and Output-Embedding Supervision}\label{sec:method}

In parallel drafting, every position predicts before its predecessors commit to tokens, so a successor cannot know which token its predecessor will take (\cref{sec:prelim-parallel}). DSpine gives each successor what its immediate predecessor is predicting. The observations in \cref{sec:obs} determine where this information comes from and when it arrives: from the adjacent predecessor, which supplies cues that earlier history cannot replace, and from the first layer on, since shallow prefix states already carry recoverable cues and accurate predecessor features help more when supplied earlier.

\begin{figure}[t]
\centering
\includegraphics[width=\linewidth]{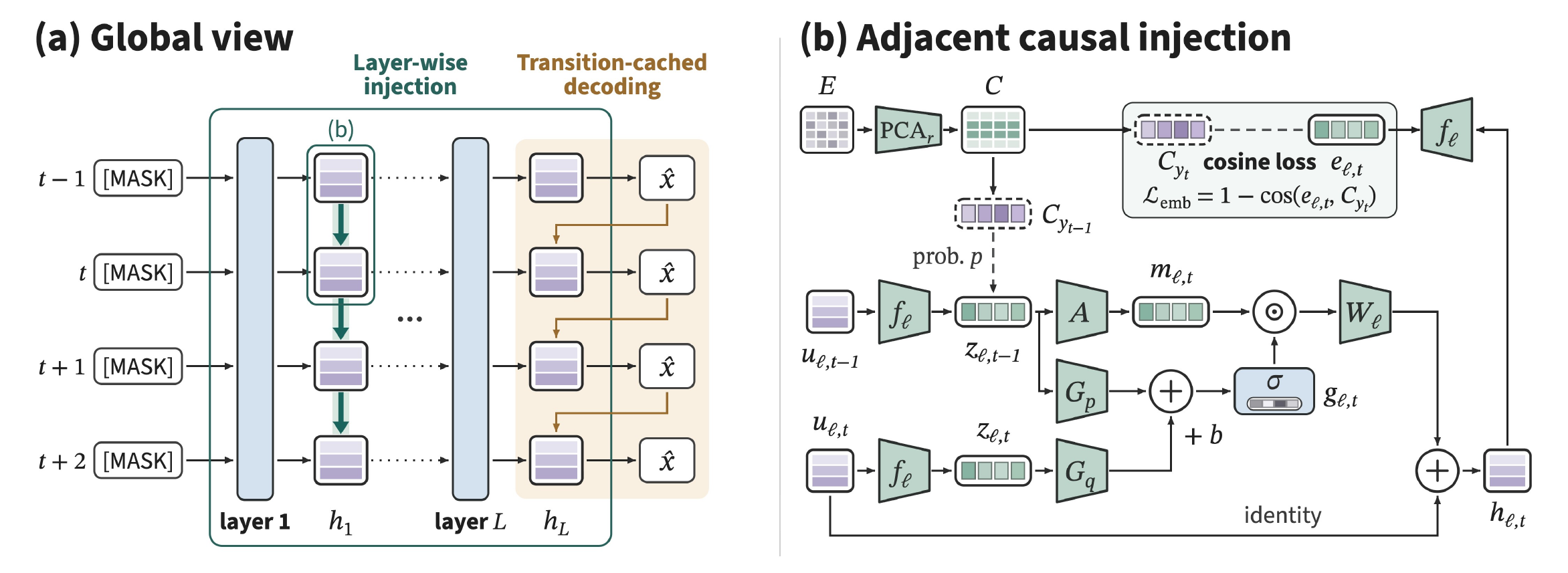}
\caption{Overview of DSpine. (a) Injection passes predecessor features to successors at every layer (green); after the final layer, transition-cached decoding selects each token conditioned on its selected predecessor (orange). (b) One injection layer: predecessor and successor features jointly gate how much of the predecessor's message enters the successor's residual stream; training adds cosine alignment to $C$ and, with probability $p$, substitutes $C_{y_{t-1}}$ for the predecessor feature.}
\label{fig:overview}
\end{figure}

\Cref{fig:overview}(a) outlines the resulting forward pass. After every Transformer layer, each position receives the feature its predecessor currently predicts, and a gate decides how much of it to absorb; all positions still update in parallel within a single forward pass. After the final layer, each position proposes its top-$K$ candidates, and tokens are selected from left to right: once a predecessor's token is selected, it is written into its successor through the same injection, which yields the successor's final scores. We describe what is passed (\cref{sec:method-space}), how it is passed inside the backbone (\cref{sec:method-injection}), how the same pathway completes decoding (\cref{sec:method-decoding}), and how the passed features are supervised and the drafter is trained (\cref{sec:method-embedding,sec:method-training}).

\subsection{Unified Transfer Space}\label{sec:method-space}

Existing parallel drafters pass predecessor information to a successor in two ways: inside the backbone, as hidden states mixed across positions by attention or similar operations; after the backbone, as the selected predecessor token fed to an additional module. The former mixes the predecessor's prediction with the context it has gathered, leaving the successor to infer which token the predecessor favors; the latter provides a decided token but only after all layers have been computed. DSpine passes the predecessor's prediction itself from the first layer on, expressed in a space where vectors name tokens.

The target LM head scores token $v$ as $E_v^\top h$, so its output embeddings naturally encode which token a state predicts. We build the unified transfer space from them: we $\ell_2$-normalize the rows of the LM head, apply PCA whitening, keep $r$ principal directions, and normalize each row again to obtain a fixed embedding table $C$ (\cref{app:config}); whitening removes the component shared by all tokens, so that directions in $C$ distinguish tokens. Two kinds of information enter this space. A predecessor whose token is still undetermined contributes a predicted feature: each layer projects its pre-injection state $u_{\ell,t}$, the output of the $\ell$-th Transformer layer, to $z_{\ell,t}=f_\ell(u_{\ell,t})=\sqrt r\,\normalize(R_\ell u_{\ell,t})$ without intermediate vocabulary decoding, where $R_\ell$ is a layer-specific learned projection to $r$ dimensions and $\normalize$ denotes $\ell_2$ normalization. A known token $x$, namely the anchor or a decoded predecessor, enters directly as its embedding $\sqrt r\,C_x$. Because both take the same form, a single pathway can carry either a prediction or a decided token: DSpine passes predicted features inside the backbone and decoded tokens in predecessor-conditioned decoding.

\subsection{Adjacent Causal Injection}\label{sec:method-injection}\label{sec:fused-injection}

After the $\ell$-th Transformer layer, position $t$ receives a message from its immediate predecessor only. Shallow predecessor features are still forming and vary in reliability, so the successor gates how much of the message to absorb: the receiver feature $z_{\ell,t}$ and the predecessor feature $z_{\ell,t-1}$ jointly determine the gate, while the predecessor feature is projected into a message that is gated and written back to the residual stream:
\begin{align}
g_{\ell,t}&=\sigma\!\left(G_qz_{\ell,t}+G_pz_{\ell,t-1}+b\right),\label{eq:gate}\\
m_{\ell,t}&=Az_{\ell,t-1},\notag\\
h_{\ell,t}&=u_{\ell,t}+\RMS(u_{\ell,t})\,W_\ell\left(g_{\ell,t}\odot m_{\ell,t}\right).\label{eq:write}
\end{align}
Here $\sigma$ is the sigmoid function; $A$, $G_q$, $G_p$, and $b$ are shared across layers, and each layer has its own write matrix $W_\ell$. All messages use pre-update features, so injection within a layer runs in parallel across positions; the anchor remains unchanged.

\paragraph{Block-causal attention.} Unlike DFlash and other parallel drafters with bidirectional block attention, DSpine uses block-causal attention in all $L$ backbone layers: position $t$ attends only to target-context K/V and block positions $0,\ldots,t$. Attention and injection thus share the predecessor-to-successor direction, and the correct information injected into a position during training cannot flow back to earlier positions (\cref{sec:method-training}). Drafting remains a single, block-parallel forward pass.

\paragraph{Propagation through depth.} Because earlier access to accurate predecessor features benefits successor prediction (\cref{fig:predecessor_content_timing}(c)), injection starts at the first layer and its effects propagate through depth: messages received at one layer enter the next layer's computation, shaping the predicted features that successors send onward. For example, information that $t-1$ receives from $t-2$ at layer $\ell$ can influence the message sent to $t$ at layer $\ell+1$ (\cref{fig:overview}(a)). The causal conditioning chain thus unfolds over network depth while positions within each layer update in parallel; earlier context remains accessible through block-causal attention.

\paragraph{Fused layer-wise injection.} Layer-wise injection consists of small normalization, projection, gating, and residual-write operations, which a direct implementation executes as many kernel launches with intermediate tensor traffic. We merge the three projections into one matrix multiplication and use two fused kernels, one for the gated message and one for the residual write together with the next layer's input normalization, all executed within the draft backbone's CUDA graph.

\subsection{Predecessor-Conditioned Decoding}\label{sec:method-decoding}\label{sec:method-inference}\label{sec:transition-cache}

\begin{wrapfigure}{r}{0.45\textwidth}
\centering
\vspace{-10pt}
\includegraphics[width=\linewidth]{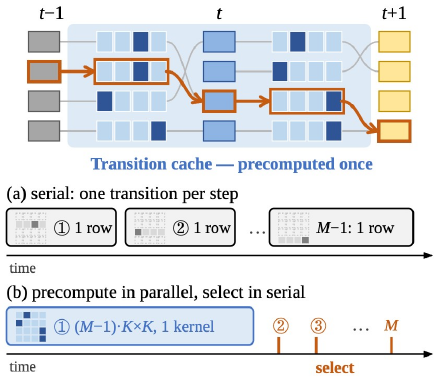}
\caption{Transition cache. (a) Serial conditional scoring. (b) Parallel transition precomputation followed by sequential selection; greedy implementation shown.}
\label{fig:transition_cache}
\vspace{-6pt}
\end{wrapfigure}
Predecessor-conditioned decoding conditions each position on the token actually decoded at its predecessor, rather than on its predicted feature, and it needs no new module. The final-layer output $h_{L,t}$ first passes through the final RMSNorm and shared LM head to determine each position's initial top-$K$ candidates. A decoded predecessor is a known token, so it enters the unified transfer space directly and passes through the same final-layer injection: for a token $s$ at position $t-1$, we replace the predicted feature $z_{L,t-1}$ with $\sqrt r\,C_s$, changing both gate and message, and redo the final-layer write (\cref{eq:write}) from the cached pre-injection state $u_{L,t}$. We denote the output by $h_t^{\mathrm{o}}(s)$ and call this step \emph{last-write refinement}; the shared LM head maps it to conditional scores $S_t(s,v)$ over the initial candidates.

\paragraph{Transition-cached decoding.} Because each position's scores depend on the predecessor's choice, computing them directly proceeds sequentially across positions. The transition cache removes this sequential computation: once the backbone has fixed each position's pre-injection state $u_{L,t}$ and top-$K$ candidate set $\mathcal V_t=\{v_{t,1},\ldots,v_{t,K}\}$, last-write refinement depends only on the predecessor's token, so before selection we compute the refined final-layer features for all predecessor candidates in parallel and score the successor candidates, organized as $T_t[i,j]=S_t(v_{t-1,i},v_{t,j})$ (\cref{fig:transition_cache}). Enumerating all $K^2$ pairs per transition trades additional scoring work for a shorter sequential dependency chain; in practice, $K=16$ suffices at a small additional cost (\cref{sec:exp-latency}). Selection then proceeds through the cache from the first position, without any sequential matrix operations.

\subsection{Layer-Wise Output-Embedding Supervision}\label{sec:method-embedding}

The predicted features that injection passes serve two roles, and the final-layer loss shapes them only indirectly through later computation. First, they condition successors from the first layer on, yet in shallow layers they carry recoverable cues but are still forming (\cref{fig:conditioning_observations}(a)). Second, the injection receives predicted features inside the backbone but token embeddings in predecessor-conditioned decoding, so the two should lie close in the unified transfer space. DSpine therefore aligns the predicted features at every layer directly with the correct token's embedding, promoting their formation in shallow layers and narrowing their gap to the token embeddings used in predecessor-conditioned decoding.

\paragraph{Directional alignment.} With the projection $f_\ell$ used for injection, we read $e_{\ell,i}=f_\ell(h_{\ell,i})$ from post-injection states and align it with the correct token's embedding $C_{y_i}$:
\begin{equation}
\cL_{\mathrm{emb}}
=\sum_{\ell=1}^{L}\sum_{i}w_i
\left[1-\cos\!\left(e_{\ell,i},C_{y_i}\right)\right].
\label{eq:emb-loss}
\end{equation}
where $w_i=\exp(-(t_i-1)/\gamma)$ is the decay weight for block position $t_i$. Because the readout follows injection, this loss trains both the formation of each feature and the successor's use of the received message; it constrains only the direction of a low-dimensional projection, leaving the full hidden state free for contextual computation.

\subsection{Training}\label{sec:method-training}

We sample draft blocks from training sequences and train the drafter with the target model, shared input embeddings, and LM head frozen. Training supervises two outputs. The first is the ordinary final-layer output, which predicts over the full vocabulary and determines the top-$K$ candidates. The second is the output of last-write refinement: as at inference, it replaces the predicted predecessor feature with a token embedding and redoes the final-layer write, except that the correct predecessor token is used, and predicts only over those candidates. Both receive the same loss below, jointly training candidate generation and predecessor-conditioned decoding.

Let $q_i$ and $p_i$ denote the draft and target distributions at position $i$ under the correct prefix. Following the loss design of DSpark \citep{cheng2026dspark}, the loss is a weighted mix of token cross-entropy and probability $\ell_1$ distance:
\begin{equation}
\cL_{\mathrm{dist}}=\sum_{i}w_i\Bigl[-\alpha\log q_i(y_i)+(1-\alpha)\,\|q_i-p_i\|_1\Bigr],
\label{eq:distribution-loss}
\end{equation}
where $\alpha$ is the mixing weight. The output of last-write refinement is also included in $\cL_{\mathrm{dist}}$, except that its distributions are restricted to the top-$K$ candidates, with the target renormalized to that set.

Early in training, predicted features carry little reliable information, which gives the injection little to learn from. We therefore replace shallow-layer predecessor features with correct-token embeddings with some probability, keeping receiver features model-derived \citep{bengio2015scheduled}, and gradually phase this replacement out; the unified transfer space makes this substitution direct. Last-write refinement always uses correct predecessors. The overall training objective weights layer-wise supervision by $\lambda$ (coefficients and schedules in \cref{app:config}):
\begin{equation}
\cL=\cL_{\mathrm{dist}}+\lambda\,\cL_{\mathrm{emb}}.
\label{eq:objective}
\end{equation}

\section{Experiments}
\label{sec:experiments}

\subsection{Experimental Setup}
\label{sec:exp-setup}

\paragraph{Models and evaluations.}
We conduct experiments on Qwen3-4B and Qwen3-8B \citep{yang2025qwen3} with thinking mode disabled, covering three task categories: \emph{Math}: GSM8K \citep{cobbe2021gsm8k} and MATH-500 \citep{lightman2023verify}; \emph{Code}: HumanEval \citep{chen2021humaneval}, MBPP \citep{austin2021mbpp}, and LiveCodeBench (LCB) \citep{jain2024livecodebench}; \emph{Chat}: MT-Bench \citep{zheng2023mtbench} and Arena-Hard \citep{li2024arenahard}. On all seven benchmarks, we measure the average acceptance length $\tau$ (\cref{sec:prelim-spec}) in SGLang \citep{zheng2023sglang} at temperatures 0 and 1; serving throughput is measured on GSM8K, MATH-500, HumanEval, and MBPP at temperature 0 and concurrency 2--32, with the same fixed verification budget for all methods (\cref{app:exp-details}).

\paragraph{Baselines.}
We compare with four representative state-of-the-art parallel drafters: DFlash \citep{chen2026dflash}, Domino \citep{huang2026domino}, DSpark \citep{cheng2026dspark}, and DFlash2 \citep{inco2026dflash2}. DFlash selects tokens independently across positions; the other three build on the DFlash backbone to model intra-block dependencies.

\paragraph{Implementation.}
For each target model, all drafters are retrained on the same ShareGPT data, with responses regenerated by the target; backbones use five layers and a block size of 16 and are trained for six epochs with the same global batch size and learning-rate schedule (\cref{app:config}).

\subsection{Main Results}
\label{sec:exp-main}

\begin{table}[t]
  \centering
  \caption{Average acceptance length $\tau$ on Qwen3 models. Each cell gives temperature 0 / 1; bold marks the best result for each model and temperature, and underline marks the second best.}
  \label{tab:main-acceptance}
  \setlength{\tabcolsep}{3pt}
  \renewcommand{\arraystretch}{1.12}
  \resizebox{\linewidth}{!}{%
    \begin{tabular}{@{}cl*{8}{c}@{}}
      \toprule
      Model & Method & \multicolumn{2}{c}{\textsc{Math}} & \multicolumn{3}{c}{\textsc{Code}} & \multicolumn{2}{c}{\textsc{Chat}} & \textsc{Overall} \\
      \cmidrule(lr){3-4}\cmidrule(lr){5-7}\cmidrule(lr){8-9}\cmidrule(lr){10-10}
       & & GSM8K & MATH-500 & HumanEval & MBPP & LCB & MT-Bench & Arena-Hard & \textit{Avg.} \\
      \midrule
        & DFlash & 4.25\,/\,3.96 & 4.23\,/\,3.86 & 3.92\,/\,3.59 & 3.93\,/\,3.56 & 3.53\,/\,2.99 & 3.27\,/\,2.97 & 3.30\,/\,2.80 & 3.77\,/\,3.39 \\
        & Domino & 4.83\,/\,4.52 & 4.81\,/\,4.30 & 4.21\,/\,3.79 & 4.31\,/\,3.85 & 3.90\,/\,3.23 & 3.58\,/\,3.18 & 3.61\,/\,2.99 & 4.18\,/\,3.69 \\
      Qwen3-8B & DSpark & 4.99\,/\,\underline{4.61} & 4.90\,/\,\underline{4.48} & \underline{4.43}\,/\,\underline{4.07} & 4.50\,/\,\underline{4.12} & 4.01\,/\,3.49 & 3.69\,/\,\underline{3.37} & 3.69\,/\,3.26 & 4.32\,/\,\underline{3.92} \\
        & DFlash2 & \underline{5.03}\,/\,4.56 & \underline{5.10}\,/\,4.47 & \underline{4.43}\,/\,4.04 & \underline{4.52}\,/\,4.06 & \underline{4.09}\,/\,\underline{3.50} & \underline{3.71}\,/\,3.35 & \underline{3.82}\,/\,\underline{3.29} & \underline{4.39}\,/\,3.90 \\
       \rowcolor{dspinerow} & DSpine (ours) & \textbf{5.67}\,/\,\textbf{5.31} & \textbf{5.49}\,/\,\textbf{5.05} & \textbf{4.94}\,/\,\textbf{4.63} & \textbf{4.98}\,/\,\textbf{4.61} & \textbf{4.46}\,/\,\textbf{3.91} & \textbf{4.11}\,/\,\textbf{3.79} & \textbf{4.12}\,/\,\textbf{3.62} & \textbf{4.82}\,/\,\textbf{4.42} \\
      \cmidrule(lr){1-10}
        & DFlash & 4.16\,/\,3.98 & 4.14\,/\,3.82 & 3.77\,/\,3.62 & 3.92\,/\,3.65 & 3.42\,/\,2.99 & 3.29\,/\,3.16 & 3.26\,/\,2.89 & 3.71\,/\,3.44 \\
        & Domino & 4.79\,/\,4.58 & 4.73\,/\,4.31 & 4.22\,/\,4.00 & 4.39\,/\,4.09 & 3.79\,/\,3.25 & 3.66\,/\,3.42 & 3.61\,/\,3.16 & 4.17\,/\,3.83 \\
      Qwen3-4B & DSpark & 5.00\,/\,\underline{4.64} & 4.90\,/\,4.47 & 4.38\,/\,\underline{4.12} & 4.59\,/\,\underline{4.20} & 3.93\,/\,3.48 & 3.80\,/\,\underline{3.51} & 3.71\,/\,3.33 & 4.33\,/\,3.96 \\
        & DFlash2 & \underline{5.14}\,/\,4.63 & \underline{5.15}\,/\,\underline{4.51} & \underline{4.59}\,/\,4.10 & \underline{4.72}\,/\,\underline{4.20} & \underline{4.10}\,/\,\underline{3.50} & \underline{3.90}\,/\,\underline{3.51} & \underline{3.91}\,/\,\underline{3.44} & \underline{4.50}\,/\,\underline{3.98} \\
       \rowcolor{dspinerow} & DSpine (ours) & \textbf{5.53}\,/\,\textbf{5.22} & \textbf{5.38}\,/\,\textbf{4.94} & \textbf{4.79}\,/\,\textbf{4.59} & \textbf{5.04}\,/\,\textbf{4.67} & \textbf{4.29}\,/\,\textbf{3.79} & \textbf{4.19}\,/\,\textbf{3.84} & \textbf{4.02}\,/\,\textbf{3.64} & \textbf{4.75}\,/\,\textbf{4.38} \\
      \bottomrule
    \end{tabular}%
  }
\end{table}

As shown in \cref{tab:main-acceptance}, DSpine achieves the longest acceptance length on all benchmarks for both models at both temperatures. On Qwen3-8B, it raises the average $\tau$ of DFlash from 3.77 to 4.82 and leads the strongest baseline by 10.0\% with greedy decoding (DFlash2) and by 12.8\% under sampling (DSpark); on Qwen3-4B, its lead over the strongest baseline, DFlash2, is 5.4\% and 10.1\%, respectively. Domino, DSpark, and DFlash2, which condition on predecessors after the backbone, already improve substantially over DFlash, while DSpine, which conditions on predecessors at every layer and again at token selection, extends the gain further, with a larger lead under sampling (see \cref{app:exp-details} for each method's proposal policy at temperature one).

\subsection{Serving Throughput}
\label{sec:exp-serving}

\begin{figure}[t]
  \centering
  \includegraphics[width=\linewidth]{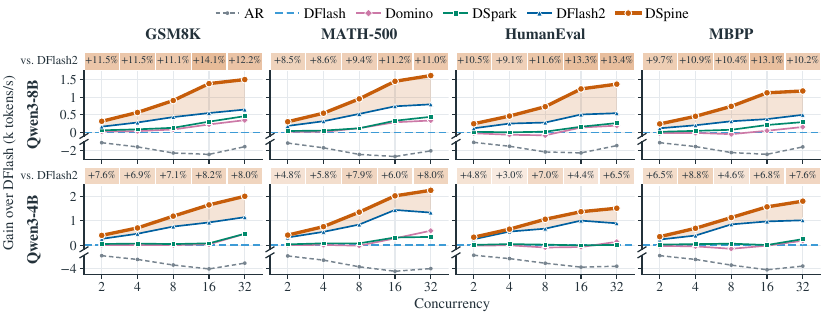}
  \caption{Serving throughput gain over DFlash on SGLang, in thousands of tokens per second (temperature zero). The dashed zero line is DFlash; shading and the strip above each panel give the lead of DSpine over DFlash2, the strongest baseline.}
  \label{fig:sglang-throughput}
\end{figure}

As shown in \cref{fig:sglang-throughput}, DSpine achieves the highest throughput at every task and concurrency level on both models: on Qwen3-8B it outperforms the strongest baseline, DFlash2, by 11.1\% on average and reaches up to 4.0$\times$ speedup over autoregressive decoding (\cref{tab:sglang-throughput}); on Qwen3-4B it leads DFlash2 by 6.5\% on average. The gain in acceptance length thus carries over to serving, and the extra cost of layer-wise injection and transition-cached decoding (\cref{sec:exp-latency}) does not offset it.

\subsection{Decoding Latency Breakdown}
\label{sec:exp-latency}

\begin{figure}[t]
  \centering
  \includegraphics[width=\linewidth]{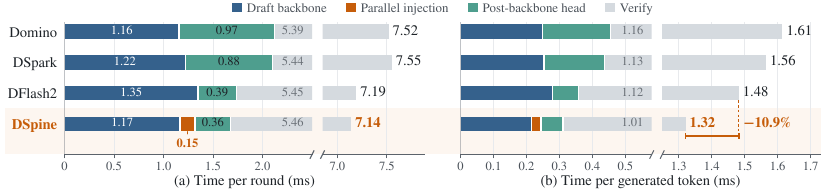}
  \caption{Module-level breakdown of Qwen3-8B decoding time on SGLang.}
  \label{fig:latency-breakdown}
\end{figure}

To examine where decoding time goes, we time each decoding round of Qwen3-8B in SGLang, split into a draft stage and a verification stage. As shown in \cref{fig:latency-breakdown}, verification takes 72--76\% of round time and costs about the same for all methods. DSpine's parallel injection adds at most 0.15\,ms per round across all five layers, and its predecessor-conditioned decoding, which scores candidates and selects through the transition cache (\cref{sec:transition-cache}), costs less than half as much as the post-backbone heads of DSpark and Domino. Overall, a DSpine round costs about as much as a DFlash2 round (7.14 vs.\ 7.19\,ms), and because more tokens are accepted per round, DSpine spends 10.9\% less decoding time per generated token than DFlash2.

\subsection{Ablation Study}
\label{sec:exp-ablation}

\begin{table}[t]
\centering
\begin{minipage}[t]{0.49\linewidth}
\centering
\caption{Ablation of module}
\label{tab:module-ablation}
\begingroup
\footnotesize
\setlength{\tabcolsep}{3pt}
\renewcommand{\arraystretch}{1.12}
\begin{tabular*}{\linewidth}{@{\extracolsep{\fill}}lccc@{}}
\toprule
Dataset & Full & \shortstack{w/o last-write\\refinement} & \shortstack{w/o refinement\\\& injection} \\
\midrule
MATH-500 & \textbf{5.49} & 5.11 & 4.93 \\
MBPP & \textbf{4.98} & 4.65 & 4.54 \\
MT-Bench & \textbf{4.11} & 3.82 & 3.67 \\
\bottomrule
\end{tabular*}
\endgroup
\end{minipage}\hfill
\begin{minipage}[t]{0.49\linewidth}
\centering
\caption{Ablation of supervision loss}
\label{tab:supervision-ablation}
\begingroup
\footnotesize
\setlength{\tabcolsep}{3pt}
\renewcommand{\arraystretch}{1.12}
\begin{tabular*}{\linewidth}{@{\extracolsep{\fill}}lccc@{}}
\toprule
Dataset & \shortstack{None\\(DFlash)} & \shortstack{Token\\CE} & \shortstack{Embedding\\cosine} \\
\midrule
MATH-500 & 3.75 & 3.67 & \textbf{3.85} \\
MBPP & 3.48 & 3.40 & \textbf{3.63} \\
MT-Bench & 2.91 & 2.84 & \textbf{2.99} \\
\bottomrule
\end{tabular*}
\endgroup
\end{minipage}
\end{table}

\paragraph{Effect of each component.}
We train each variant in \cref{tab:module-ablation} independently. Removing last-write refinement reduces $\tau$ by 6.6--7.1\%, and further removing parallel injection enlarges the drop to 8.8--10.7\%. Both components improve acceptance length, and their contributions are complementary.

\paragraph{Supervision target.}
\Cref{tab:supervision-ablation} compares intermediate-layer supervision: directional supervision toward compact output embeddings improves acceptance length, whereas per-layer full-vocabulary cross-entropy falls below the unsupervised baseline. This supports the design of \cref{sec:method-embedding}, as forcing intermediate states to classify tokens may interfere with later layers' contextual computation.

\section{Related Work}
\label{sec:related}

Speculative decoding lets a lightweight drafter propose tokens that the target model verifies losslessly \citep{leviathan2023fast,chen2023accelerating}. Follow-ups mainly improve the drafter: Medusa predicts several positions with independent heads, Hydra makes these heads depend on drafted tokens \citep{cai2024medusa,ankner2024hydra}, the EAGLE series drafts autoregressively from target features \citep{li2024eagle,li2024eagle2,li2025eagle3}, and tree verification checks multiple continuations per round \citep{miao2024specinfer,li2024eagle2}. Autoregressive drafting conditions each token on its predecessors but costs one sequential step per token.

Parallel decoding methods predict multiple positions at once \citep{stern2018blockwise,an2026pard,liu2026dart,christopher2025speculative,li2026diffuspec,chen2025aspd}. DFlash drafts a block in one pass of a lightweight block-diffusion model conditioned on target features \citep{chen2026dflash}; follow-ups add layer-wise target conditioning, position-weighted losses, or draft trees \citep{zhang2026dflare,wu2026dpace,ringel2026ddtree}. Since positions predict before their predecessors commit, the resulting tokens can be inconsistent \citep{stern2018blockwise,cheng2026dspark}. Most remedies restore this dependency after the backbone, through a causal correction branch (Domino), a sequential head (DSpark), or a path selector over top-$k$ candidates (DFlash2) \citep{huang2026domino,cheng2026dspark,inco2026dflash2}, as well as previous-token conditioning, block refinement, or adjacent-candidate scoring \citep{rheinboldt2026treeflash,wang2026xpress,rusanovsky2026lilicorr}. Inside the backbone, DFlash2 adds causal two-tap convolutions, and DART and JetSpec use causal block attention \citep{inco2026dflash2,liu2026dart,hu2026jetspec}. DSpine also uses block-causal attention but additionally injects each predecessor's predicted feature, supervised in a unified transfer space, into its successor at every layer; the same injection carries decoded tokens in predecessor-conditioned decoding.

\section{Conclusion}
\label{sec:conclusion}

DSpine injects causal conditioning throughout the backbone, building intra-block token dependencies over network depth within a single parallel drafting pass. Gated adjacent injection passes predecessor features to successors at every layer, and a unified transfer space with layer-wise supervision links this injection to predecessor-conditioned decoding. A transition cache precomputes the scores of all adjacent candidate pairs in parallel, turning sequential conditional decoding into parallel computation. Across math, code, and chat tasks, DSpine improves acceptance length over representative parallel drafters and translates these gains into higher SGLang serving throughput. These results motivate further study of in-layer information transfer, including more flexible communication ranges and interactions in other parallel decoding architectures.

\bibliography{references}

\begin{thebibliography}{36}
\providecommand{\natexlab}[1]{#1}
\providecommand{\url}[1]{\texttt{#1}}
\expandafter\ifx\csname urlstyle\endcsname\relax
  \providecommand{\doi}[1]{doi: #1}\else
  \providecommand{\doi}{doi: \begingroup \urlstyle{rm}\Url}\fi

\bibitem[Leviathan et~al.(2023)Leviathan, Kalman, and
  Matias]{leviathan2023fast}
Yaniv Leviathan, Matan Kalman, and Yossi Matias.
\newblock Fast inference from transformers via speculative decoding.
\newblock In Andreas Krause, Emma Brunskill, Kyunghyun Cho, Barbara Engelhardt,
  Sivan Sabato, and Jonathan Scarlett, editors, \emph{Proceedings of the 40th
  International Conference on Machine Learning}, volume 202 of
  \emph{Proceedings of Machine Learning Research}, pages 19274--19286. PMLR,
  2023.

\bibitem[Li et~al.(2025)Li, Wei, Zhang, and Zhang]{li2025eagle3}
Yuhui Li, Fangyun Wei, Chao Zhang, and Hongyang Zhang.
\newblock {EAGLE-3}: Scaling up inference acceleration of large language models
  via training-time test.
\newblock In \emph{Advances in Neural Information Processing Systems},
  volume~38, 2025.
\newblock \doi{10.52202/085713-4562}.
\newblock URL
  \url{https://papers.nips.cc/paper_files/paper/2025/hash/c7b5a35ea98b62512a869c19ea7b03cb-Abstract-Conference.html}.

\bibitem[Chen et~al.(2026)Chen, Liang, and Liu]{chen2026dflash}
Jian Chen, Yesheng Liang, and Zhijian Liu.
\newblock {DFlash}: Block diffusion for flash speculative decoding.
\newblock In \emph{Proceedings of the 43rd International Conference on Machine
  Learning}, volume 306 of \emph{Proceedings of Machine Learning Research}.
  PMLR, 2026.

\bibitem[{DFlash2 Team}(2026)]{inco2026dflash2}
{DFlash2 Team}.
\newblock {DFlash 2}: Keep drafting parallel.
\newblock Blog post, August 2026.
\newblock URL \url{https://inco.ai/blog/dflash2/}.

\bibitem[Huang et~al.(2026)Huang, Zhang, Zhang, Lin, Xu, and
  Zhang]{huang2026domino}
Jianuo Huang, Yaojie Zhang, Qituan Zhang, Hao Lin, Hanlin Xu, and Linfeng
  Zhang.
\newblock {Domino}: Decoupling causal modeling from autoregressive drafting in
  speculative decoding.
\newblock arXiv preprint arXiv:2605.29707, 2026.
\newblock URL \url{https://arxiv.org/abs/2605.29707}.

\bibitem[Cheng et~al.(2026)Cheng, Yu, Shao, Li, Xiong, Qian, Zhu, Ma, Zhang,
  Ye, Chen, Deng, Yu, Dai, Zhang, Wei, Tan, Yang, Xu, Wu, Xu, Wang, Chen, Tian,
  Bi, Hao, Chen, Cao, Zhang, Xu, Zhang, Zhao, and Liang]{cheng2026dspark}
Xin Cheng, Xingkai Yu, Chenze Shao, Jiashi Li, Yunfan Xiong, Yi~Qian, Jiaqi
  Zhu, Shirong Ma, Xiaokang Zhang, Jiasheng Ye, Qinyu Chen, Chengqi Deng,
  Jiping Yu, Damai Dai, Zhengyan Zhang, Yixuan Wei, Yixuan Tan, Wenkai Yang,
  Runxin Xu, Yu~Wu, Zhean Xu, Xuanyu Wang, Muyang Chen, Rui Tian, Xiao Bi,
  Zhewen Hao, Shaoyuan Chen, Huanqi Cao, Wentao Zhang, Anyi Xu, Huishuai Zhang,
  Dongyan Zhao, and Wenfeng Liang.
\newblock {DSpark}: Confidence-scheduled speculative decoding with
  semi-autoregressive generation.
\newblock arXiv preprint arXiv:2607.05147, 2026.
\newblock URL \url{https://arxiv.org/abs/2607.05147}.

\bibitem[Zhang et~al.(2026)Zhang, Yu, Liu, Yu, Li, Zhu, Duo, Xiong, Song, Yu,
  Zhu, and Li]{zhang2026dflare}
Jiebin Zhang, Zhenghan Yu, Song Liu, Eugene~J. Yu, Zheng Li, Dawei Zhu,
  Jiangshan Duo, Weimin Xiong, Yifan Song, Guanghua Yu, Jianchen Zhu, and
  Sujian Li.
\newblock {DFlare}: Scaling up draft capacity for block diffusion speculative
  decoding.
\newblock arXiv preprint arXiv:2606.02091, 2026.
\newblock URL \url{https://arxiv.org/abs/2606.02091}.

\bibitem[Bengio et~al.(2015)Bengio, Vinyals, Jaitly, and
  Shazeer]{bengio2015scheduled}
Samy Bengio, Oriol Vinyals, Navdeep Jaitly, and Noam Shazeer.
\newblock Scheduled sampling for sequence prediction with recurrent neural
  networks.
\newblock In C.~Cortes, N.~Lawrence, D.~Lee, M.~Sugiyama, and R.~Garnett,
  editors, \emph{Advances in Neural Information Processing Systems}, volume~28.
  Curran Associates, Inc., 2015.

\bibitem[Yang et~al.(2025)Yang, Li, Yang, Zhang, Hui, Zheng, Yu, Gao, Huang,
  Lv, et~al.]{yang2025qwen3}
An~Yang, Anfeng Li, Baosong Yang, Beichen Zhang, Binyuan Hui, Bo~Zheng, Bowen
  Yu, Chang Gao, Chengen Huang, Chenxu Lv, et~al.
\newblock {Qwen3} technical report, 2025.

\bibitem[Cobbe et~al.(2021)Cobbe, Kosaraju, Bavarian, Chen, Jun, Kaiser,
  Plappert, Tworek, Hilton, Nakano, Hesse, and Schulman]{cobbe2021gsm8k}
Karl Cobbe, Vineet Kosaraju, Mohammad Bavarian, Mark Chen, Heewoo Jun, Lukasz
  Kaiser, Matthias Plappert, Jerry Tworek, Jacob Hilton, Reiichiro Nakano,
  Christopher Hesse, and John Schulman.
\newblock Training verifiers to solve math word problems, 2021.

\bibitem[Lightman et~al.(2023)Lightman, Kosaraju, Burda, Edwards, Baker, Lee,
  Leike, Schulman, Sutskever, and Cobbe]{lightman2023verify}
Hunter Lightman, Vineet Kosaraju, Yura Burda, Harri Edwards, Bowen Baker, Teddy
  Lee, Jan Leike, John Schulman, Ilya Sutskever, and Karl Cobbe.
\newblock Let's verify step by step, 2023.

\bibitem[Chen et~al.(2021)Chen, Tworek, Jun, Yuan, Pinto, Kaplan, Edwards,
  Burda, Joseph, Brockman, et~al.]{chen2021humaneval}
Mark Chen, Jerry Tworek, Heewoo Jun, Qiming Yuan, Henrique Ponde de~Oliveira
  Pinto, Jared Kaplan, Harri Edwards, Yuri Burda, Nicholas Joseph, Greg
  Brockman, et~al.
\newblock Evaluating large language models trained on code, 2021.

\bibitem[Austin et~al.(2021)Austin, Odena, Nye, Bosma, Michalewski, Dohan,
  Jiang, Cai, Terry, Le, and Sutton]{austin2021mbpp}
Jacob Austin, Augustus Odena, Maxwell Nye, Maarten Bosma, Henryk Michalewski,
  David Dohan, Ellen Jiang, Carrie Cai, Michael Terry, Quoc Le, and Charles
  Sutton.
\newblock Program synthesis with large language models, 2021.

\bibitem[Jain et~al.(2024)Jain, Han, Gu, Li, Yan, Zhang, Wang, Solar-Lezama,
  Sen, and Stoica]{jain2024livecodebench}
Naman Jain, King Han, Alex Gu, Wen-Ding Li, Fanjia Yan, Tianjun Zhang, Sida
  Wang, Armando Solar-Lezama, Koushik Sen, and Ion Stoica.
\newblock {LiveCodeBench}: Holistic and contamination free evaluation of large
  language models for code, 2024.

\bibitem[Zheng et~al.(2023{\natexlab{a}})Zheng, Chiang, Sheng, Zhuang, Wu,
  Zhuang, Lin, Li, Li, Xing, et~al.]{zheng2023mtbench}
Lianmin Zheng, Wei-Lin Chiang, Ying Sheng, Siyuan Zhuang, Zhanghao Wu, Yonghao
  Zhuang, Zi~Lin, Zhuohan Li, Dacheng Li, Eric~P. Xing, et~al.
\newblock Judging {LLM}-as-a-judge with {MT-Bench} and {Chatbot Arena},
  2023{\natexlab{a}}.

\bibitem[Li et~al.(2024{\natexlab{a}})Li, Chiang, Frick, Dunlap, Wu, Zhu,
  Gonzalez, and Stoica]{li2024arenahard}
Tianle Li, Wei-Lin Chiang, Evan Frick, Lisa Dunlap, Tianhao Wu, Banghua Zhu,
  Joseph~E. Gonzalez, and Ion Stoica.
\newblock From crowdsourced data to high-quality benchmarks: {Arena-Hard} and
  {BenchBuilder} pipeline, 2024{\natexlab{a}}.

\bibitem[Zheng et~al.(2023{\natexlab{b}})Zheng, Yin, Xie, Sun, Huang, Yu, Cao,
  Kozyrakis, Stoica, Gonzalez, Barrett, and Sheng]{zheng2023sglang}
Lianmin Zheng, Liangsheng Yin, Zhiqiang Xie, Chuyue Sun, Jeff Huang, Cody~Hao
  Yu, Shiyi Cao, Christos Kozyrakis, Ion Stoica, Joseph~E. Gonzalez, Clark
  Barrett, and Ying Sheng.
\newblock {SGLang}: Efficient execution of structured language model programs,
  2023{\natexlab{b}}.

\bibitem[Chen et~al.(2023)Chen, Borgeaud, Irving, Lespiau, Sifre, and
  Jumper]{chen2023accelerating}
Charlie Chen, Sebastian Borgeaud, Geoffrey Irving, Jean-Baptiste Lespiau,
  Laurent Sifre, and John Jumper.
\newblock Accelerating large language model decoding with speculative sampling.
\newblock arXiv preprint arXiv:2302.01318, 2023.
\newblock URL \url{https://arxiv.org/abs/2302.01318}.

\bibitem[Cai et~al.(2024)Cai, Li, Geng, Peng, Lee, Chen, and
  Dao]{cai2024medusa}
Tianle Cai, Yuhong Li, Zhengyang Geng, Hongwu Peng, Jason~D. Lee, Deming Chen,
  and Tri Dao.
\newblock Medusa: Simple {LLM} inference acceleration framework with multiple
  decoding heads.
\newblock In Ruslan Salakhutdinov, Zico Kolter, Katherine Heller, Adrian
  Weller, Nuria Oliver, Jonathan Scarlett, and Felix Berkenkamp, editors,
  \emph{Proceedings of the 41st International Conference on Machine Learning},
  volume 235 of \emph{Proceedings of Machine Learning Research}, pages
  5209--5235. PMLR, 2024.

\bibitem[Ankner et~al.(2024)Ankner, Parthasarathy, Nrusimha, Rinard,
  Ragan-Kelley, and Brandon]{ankner2024hydra}
Zachary Ankner, Rishab Parthasarathy, Aniruddha Nrusimha, Christopher Rinard,
  Jonathan Ragan-Kelley, and William Brandon.
\newblock Hydra: Sequentially-dependent draft heads for {Medusa} decoding.
\newblock In \emph{First Conference on Language Modeling}, 2024.
\newblock URL \url{https://openreview.net/forum?id=FbhjirzvJG}.

\bibitem[Li et~al.(2024{\natexlab{b}})Li, Wei, Zhang, and Zhang]{li2024eagle}
Yuhui Li, Fangyun Wei, Chao Zhang, and Hongyang Zhang.
\newblock {EAGLE}: Speculative sampling requires rethinking feature
  uncertainty.
\newblock In Ruslan Salakhutdinov, Zico Kolter, Katherine Heller, Adrian
  Weller, Nuria Oliver, Jonathan Scarlett, and Felix Berkenkamp, editors,
  \emph{Proceedings of the 41st International Conference on Machine Learning},
  volume 235 of \emph{Proceedings of Machine Learning Research}, pages
  28935--28948. PMLR, 2024{\natexlab{b}}.

\bibitem[Li et~al.(2024{\natexlab{c}})Li, Wei, Zhang, and Zhang]{li2024eagle2}
Yuhui Li, Fangyun Wei, Chao Zhang, and Hongyang Zhang.
\newblock {EAGLE}-2: Faster inference of language models with dynamic draft
  trees.
\newblock In Yaser Al-Onaizan, Mohit Bansal, and Yun-Nung Chen, editors,
  \emph{Proceedings of the 2024 Conference on Empirical Methods in Natural
  Language Processing}, pages 7421--7432, Miami, Florida, USA, November
  2024{\natexlab{c}}. Association for Computational Linguistics.
\newblock \doi{10.18653/v1/2024.emnlp-main.422}.
\newblock URL \url{https://aclanthology.org/2024.emnlp-main.422/}.

\bibitem[Miao et~al.(2024)Miao, Oliaro, Zhang, Cheng, Wang, Zhang, Wong, Zhu,
  Yang, Shi, Shi, Chen, Arfeen, Abhyankar, and Jia]{miao2024specinfer}
Xupeng Miao, Gabriele Oliaro, Zhihao Zhang, Xinhao Cheng, Zeyu Wang, Zhengxin
  Zhang, Rae Ying~Yee Wong, Alan Zhu, Lijie Yang, Xiaoxiang Shi, Chunan Shi,
  Zhuoming Chen, Daiyaan Arfeen, Reyna Abhyankar, and Zhihao Jia.
\newblock {SpecInfer}: Accelerating large language model serving with
  tree-based speculative inference and verification.
\newblock In \emph{Proceedings of the 29th ACM International Conference on
  Architectural Support for Programming Languages and Operating Systems, Volume
  3}, ASPLOS '24, pages 932--949. ACM, 2024.
\newblock \doi{10.1145/3620666.3651335}.

\bibitem[Stern et~al.(2018)Stern, Shazeer, and Uszkoreit]{stern2018blockwise}
Mitchell Stern, Noam Shazeer, and Jakob Uszkoreit.
\newblock Blockwise parallel decoding for deep autoregressive models.
\newblock In S.~Bengio, H.~Wallach, H.~Larochelle, K.~Grauman, N.~Cesa-Bianchi,
  and R.~Garnett, editors, \emph{Advances in Neural Information Processing
  Systems}, volume~31. Curran Associates, Inc., 2018.

\bibitem[An et~al.(2026)An, Bai, Liu, Li, and Barsoum]{an2026pard}
Zihao An, Huajun Bai, Ziqiong Liu, Dong Li, and Emad Barsoum.
\newblock {PARD}: Accelerating {LLM} inference with low-cost {PAR}allel draft
  model adaptation.
\newblock In \emph{The Fourteenth International Conference on Learning
  Representations}, 2026.
\newblock URL \url{https://openreview.net/forum?id=XbOyv7iVGL}.

\bibitem[Liu et~al.(2026)Liu, Li, Zhao, Gao, Zhou, Zhang, Wang, Dou, Zhong, and
  Tian]{liu2026dart}
Fuliang Liu, Xue Li, Ketai Zhao, Yinxi Gao, Ziyan Zhou, Zhonghui Zhang, Zhibin
  Wang, Wanchun Dou, Sheng Zhong, and Chen Tian.
\newblock {DART}: Diffusion-inspired speculative decoding for fast {LLM}
  inference.
\newblock arXiv preprint arXiv:2601.19278, 2026.
\newblock URL \url{https://arxiv.org/abs/2601.19278}.

\bibitem[Christopher et~al.(2025)Christopher, Bartoldson, Ben-Nun, Cardei,
  Kailkhura, and Fioretto]{christopher2025speculative}
Jacob~K Christopher, Brian~R. Bartoldson, Tal Ben-Nun, Michael Cardei, Bhavya
  Kailkhura, and Ferdinando Fioretto.
\newblock Speculative diffusion decoding: Accelerating language generation
  through diffusion.
\newblock In Luis Chiruzzo, Alan Ritter, and Lu~Wang, editors,
  \emph{Proceedings of the 2025 Conference of the Nations of the Americas
  Chapter of the Association for Computational Linguistics: Human Language
  Technologies (Volume 1: Long Papers)}, pages 12042--12059, Albuquerque, New
  Mexico, April 2025. Association for Computational Linguistics.
\newblock \doi{10.18653/v1/2025.naacl-long.601}.
\newblock URL \url{https://aclanthology.org/2025.naacl-long.601/}.

\bibitem[Li et~al.(2026)Li, Fu, Fang, Zhao, Tang, Yuan, and
  Wang]{li2026diffuspec}
Guanghao Li, Zhihui Fu, Min Fang, Qibin Zhao, Ming Tang, Chun Yuan, and Jun
  Wang.
\newblock {DiffuSpec}: Unlocking diffusion language models for speculative
  decoding.
\newblock In Maria Liakata, Viviane~P. Moreira, Jiajun Zhang, and David
  Jurgens, editors, \emph{Findings of the Association for Computational
  Linguistics: ACL 2026}, pages 20896--20910, San Diego, California, United
  States, July 2026. Association for Computational Linguistics.
\newblock \doi{10.18653/v1/2026.findings-acl.1048}.
\newblock URL \url{https://aclanthology.org/2026.findings-acl.1048/}.

\bibitem[Chen et~al.(2025)Chen, Shen, Yu, Wu, Wen, He, Qiao, and
  Sun]{chen2025aspd}
Keyu Chen, Zhifeng Shen, Daohai Yu, Haoqian Wu, Wei Wen, Jianfeng He, Ruizhi
  Qiao, and Xing Sun.
\newblock {ASPD}: Unlocking adaptive serial-parallel decoding by exploring
  intrinsic parallelism in {LLMs}.
\newblock arXiv preprint arXiv:2508.08895, 2025.
\newblock URL \url{https://arxiv.org/abs/2508.08895}.

\bibitem[Wu et~al.(2026)Wu, Yao, Qi, Zheng, Wang, Ma, Liao, Lakkaraju, Li, and
  Du]{wu2026dpace}
Tianyu Wu, Yu~Yao, Zhenting Qi, Han Zheng, Zhuohan Wang, Haoran Ma, Lawrence
  Liao, Himabindu Lakkaraju, Ju~Li, and Yilun Du.
\newblock {D-PACE}: Dynamic position-aware cross-entropy for parallel
  speculative drafting.
\newblock arXiv preprint arXiv:2605.18810, 2026.
\newblock URL \url{https://arxiv.org/abs/2605.18810}.

\bibitem[Ringel and Romano(2026)]{ringel2026ddtree}
Liran Ringel and Yaniv Romano.
\newblock Accelerating speculative decoding with block diffusion draft trees.
\newblock arXiv preprint arXiv:2604.12989, 2026.
\newblock URL \url{https://arxiv.org/abs/2604.12989}.

\bibitem[Rheinboldt et~al.(2026)Rheinboldt, Berdoz, and
  Wattenhofer]{rheinboldt2026treeflash}
Peer Rheinboldt, Fr{\'e}d{\'e}ric Berdoz, and Roger Wattenhofer.
\newblock {TreeFlash}: Parallel {AR}-approximation for faster speculative
  decoding.
\newblock arXiv preprint arXiv:2606.03819, 2026.
\newblock URL \url{https://arxiv.org/abs/2606.03819}.

\bibitem[Wang et~al.(2026)Wang, Wertheimer, Lim, Srivatsa, Ganti, Zhang, and
  Wang]{wang2026xpress}
Zheng Wang, Davis Wertheimer, Yu~Chin~Fabian Lim, Mudhakar Srivatsa, Raghu~K.
  Ganti, Minjia Zhang, and Naigang Wang.
\newblock {xPress}: Parallel refinement for diffusion drafters in speculative
  decoding.
\newblock arXiv preprint arXiv:2608.02438, 2026.
\newblock URL \url{https://arxiv.org/abs/2608.02438}.

\bibitem[Rusanovsky et~al.(2026)Rusanovsky, Miron, Uziel, Belhasin, Guo,
  Zilberstein, Ashkenazi, and Elad]{rusanovsky2026lilicorr}
Matan Rusanovsky, Yoav Miron, Roy Uziel, Omer Belhasin, Hao Guo, Ran
  Zilberstein, Maor Ashkenazi, and Michael Elad.
\newblock {LiLiCorr}: Lightweight likelihood correlation of parallel drafts for
  speculative decoding.
\newblock arXiv preprint arXiv:2608.20530, 2026.
\newblock URL \url{https://arxiv.org/abs/2608.20530}.

\bibitem[Hu et~al.(2026)Hu, Feng, Wu, Yuan, Zhao, Qian, Wang, Zhao, Jiang, Zhu,
  Rosing, and Zhang]{hu2026jetspec}
Lanxiang Hu, Zhaoxiang Feng, Yulun Wu, Haoran Yuan, Yujie Zhao, Yu-Yang Qian,
  Bojun Wang, Peng Zhao, Daxin Jiang, Yibo Zhu, Tajana Rosing, and Hao Zhang.
\newblock {JetSpec}: Breaking the scaling ceiling of speculative decoding with
  parallel tree drafting.
\newblock arXiv preprint arXiv:2606.18394, 2026.
\newblock URL \url{https://arxiv.org/abs/2606.18394}.

\bibitem[Shah et~al.(2024)Shah, Bikshandi, Zhang, Thakkar, Ramani, and
  Dao]{shah2024flashattention3}
Jay Shah, Ganesh Bikshandi, Ying Zhang, Vijay Thakkar, Pradeep Ramani, and Tri
  Dao.
\newblock {FlashAttention-3}: Fast and accurate attention with asynchrony and
  low-precision, 2024.

\end{thebibliography}

\newpage
\appendix
\section{Implementation Details of DSpine}
\label{app:config}

\paragraph{Architecture.}
DSpine builds on the DFlash drafter for Qwen3-8B (\cref{tab:hparams}): five Transformer layers with block-causal attention, target context features fused from five target layers and supplied as K/V to every layer, and the target's input embedding and LM head shared and frozen. Injection adds about 33M parameters, roughly 3\% of the backbone.

\paragraph{Unified transfer space.}
Let $\bar E_v$ be the $\ell_2$-normalized LM-head row of token $v$ and $\mu$ its mean over the vocabulary, and let $P_r$ and $\Lambda_r$ hold the top $r$ eigenvectors and eigenvalues of the centered Gram matrix $\sum_v(\bar E_v-\mu)(\bar E_v-\mu)^\top$. The table is $C_v=\normalize\bigl(\Lambda_r^{-1/2}P_r^\top(\bar E_v-\mu)\bigr)$, fixed and shared by all layers.

\paragraph{Initialization.}
Each readout matrix $R_\ell$ is initialized with $P_r^\top$; the write matrices $W_\ell$ and the predecessor gate matrix $G_p$ are zero-initialized with $b=0$, so every injection starts as the identity map.

\paragraph{Losses.}
Both the backbone output and the last-write refinement output use $\alpha=0.1$ in \cref{eq:distribution-loss} and are summed with equal weight. The layer-wise supervision weight is $\lambda=\beta_t\,\sg(\mathrm{CE}_{\mathrm{bb}})$, where $\mathrm{CE}_{\mathrm{bb}}$ is the token cross-entropy of the backbone output and $\beta_t$ rises linearly from 0 to 0.5 over the first 146 steps. For last-write refinement, the cross-entropy term is computed only where the correct token lies among the $K=16$ candidates, and the $\ell_1$ term only where the candidates hold more than $10^{-4}$ of the target probability, against the target distribution renormalized to the candidates.

\paragraph{Correct-predecessor curriculum.}
Each draft block is selected with probability $p$ by one draw shared by layers 1--3, which then use $\sqrt r\,C_{y_{t-1}}$ as predecessor input. $p$ is $0.5$ for the first sixth of training and decays linearly to zero at one third.

\paragraph{Training setup.}
All drafters share the same training configuration: the same ShareGPT data with responses regenerated by the target model with thinking disabled, the same draft block of one anchor and 15 candidates, a global batch of 112 sequences, AdamW without weight decay at a peak learning rate of $6\times10^{-4}$ with 4\% linear warmup (146 steps) and cosine decay to zero, a maximum training sequence length of 3{,}072 tokens, and six epochs (3{,}655 steps). \Cref{tab:hparams} lists the remaining DSpine hyperparameters.

\begin{table}[h]
\centering
\small
\caption{Remaining DSpine hyperparameters for Qwen3-8B.}
\label{tab:hparams}
\vspace{4pt}
\setlength{\tabcolsep}{10pt}
\begin{tabular}{@{}lc@{}}
\toprule
Hyperparameter & Value \\
\midrule
Draft layers $L$ & 5 \\
Hidden size & 4096 \\
Target feature layers & $\{1,9,17,25,33\}$ \\
Transfer-space dimension $r$ & 1024 \\
Message dimension $a$ & 512 \\
Candidates $K$ & 16 \\
\midrule
Anchors per sequence & 512 \\
Gradient-norm clip & 1.0 \\
Position decay $\gamma$ & 7 \\
Loss mixing $\alpha$ & 0.1 \\
\bottomrule
\end{tabular}
\end{table}

\section{Details of the Conditioning Experiments}
\label{app:obs}

\subsection{Setup, Readouts, and Metrics}
\label{app:obs-setup}
We use the GSM8K test set and HumanEval with the frozen official Qwen3-8B-DFlash-b16 drafter; greedy Qwen3-8B continuations (thinking disabled, at most 1,024 new tokens) serve as correct tokens. Each block contains one anchor and 15 candidates.

For layer $\ell$, the readout is
\begin{equation}
q_\ell=\softmax\!\left(E[n(h_\ell)+U_\ell V_\ell n(h_\ell)+b_\ell]\right),
\end{equation}
where $E$ is the frozen target LM head, $n$ is RMS normalization without a trainable scale, and $U_\ell V_\ell$ has rank 64. Each readout is fitted with cross-entropy on 500 GSM8K training questions and selected on another 100, both disjoint from the test data (AdamW, three epochs, batch size 256, learning rate $10^{-3}$, weight decay $10^{-4}$, 5\% warmup, cosine decay).

For $T$ evaluated positions, the number of consecutive correct tokens is
\begin{equation}
K_T=\sum_{j=1}^{T}\prod_{t=1}^{j}\mathbb{I}[\hat y_t=y_t],
\end{equation}
which, unlike $\tau$, compares predictions with fixed correct continuations. We average blocks within each task and then over tasks; panels (b) and (c) of \cref{fig:conditioning_observations} average GSM8K and HumanEval with equal weight.

\subsection{Prefix Content and Timing}
\label{app:obs-prefix}
We fix the correct token $x_7$ as a new anchor and fill $x_1,\ldots,x_6$ with either draft predictions (P) or correct tokens (G), keeping earlier context unchanged. The target model encodes the context up to $x_6$, and its fused features enter each draft layer as K/V; layer 1 always receives P features, while layers 2--5 follow the chosen schedule. We evaluate $x_8,\ldots,x_{15}$ and report the relative gain in $K_8$ over the all-P schedule. \Cref{tab:prefix-schedules} lists all schedules, and \cref{fig:prefix_content_timing}(b) shows the average of the two datasets.

\begin{table}[t]
\centering
\small
\caption{Prefix-timing conditions. Schedules list layers 1--5; P/G denotes predicted/correct prefix features. Gains in $K_8$ are relative to the all-P baseline.}
\label{tab:prefix-schedules}
\begin{tabular}{lrr}
\toprule
Schedule & GSM8K gain (\%) & HumanEval gain (\%)\\
\midrule
P P P P P & +0.00 & +0.00\\
P G G G G & +6.23 & +7.12\\
P P G G G & +1.87 & +3.93\\
P P P P G & $-$0.18 & +0.26\\
P G P P G & +4.18 & +4.50\\
P P G P G & +1.06 & +3.09\\
P P P G G & +0.70 & +1.73\\
\bottomrule
\end{tabular}
\end{table}

\subsection{Adjacent Predecessor versus Earlier History}
\label{app:obs-predecessor}
For original-block positions 4, 8, and 12, we use the immediately preceding token as a new anchor; the earlier history and the anchor independently take draft predictions or correct tokens. The history is encoded by the target model and the predecessor enters the drafter through its input embedding, so each evaluated token is the first candidate of its new block. \Cref{tab:predecessor-content} reports the gain from correcting only the predecessor; the L1--L5 points in \cref{fig:predecessor_content_timing}(c) average it over the three positions. Correcting the predecessor helps in every setting; at position 12 with correct history, accuracy rises from 51.00\% to 93.50\% on GSM8K and from 48.78\% to 91.31\% on HumanEval.

\begin{table}[t]
\centering
\small
\caption{Accuracy gain (percentage points) from correcting only the predecessor. History refers to tokens before the predecessor; positions refer to the original block.}
\label{tab:predecessor-content}
\begin{tabular}{llrrr}
\toprule
Dataset & History & Position 4 & Position 8 & Position 12\\
\midrule
GSM8K & Predicted & 14.88 & 19.63 & 23.38\\
GSM8K & Correct & 17.62 & 31.87 & 42.50\\
HumanEval & Predicted & 14.02 & 23.63 & 19.51\\
HumanEval & Correct & 17.53 & 38.72 & 42.53\\
\bottomrule
\end{tabular}
\end{table}

\subsection{Timing of Predecessor Features}
\label{app:obs-timing}
We cache the predecessor's K/V at every layer in two forward passes, one with the predicted and one with the correct predecessor, and substitute the correct K/V only at selected layers. The comparison supplies correct features twice: at layer 5 and at one of layers 2--4. The all-predicted and all-correct schedules reproduce the two native forward passes, so the L1--L5 points in \cref{fig:predecessor_content_timing}(c) equal the content effect in \cref{app:obs-predecessor}. Averaged over the three positions, L2+L5 improves accuracy over L4+L5 by 14.33 and 15.24 points on GSM8K and HumanEval with correct history (\cref{tab:predecessor-timing}), and by 9.63 and 8.99 points with predicted history.

\begin{table}[t]
\centering
\small
\caption{L2+L5 minus L4+L5 with correct earlier history. Accuracy differences are in percentage points; $K_4$ counts at most four consecutive correct candidates, excluding the anchor.}
\label{tab:predecessor-timing}
\begin{tabular}{llrr}
\toprule
Dataset & Position & Accuracy difference & $K_4$ difference\\
\midrule
GSM8K & 4 & 7.38 & 0.226\\
GSM8K & 8 & 17.00 & 0.565\\
GSM8K & 12 & 18.63 & 0.755\\
HumanEval & 4 & 11.28 & 0.280\\
HumanEval & 8 & 15.85 & 0.694\\
HumanEval & 12 & 18.60 & 0.745\\
\bottomrule
\end{tabular}
\end{table}

\section{Experimental Details}
\label{app:exp-details}

\setlength{\textfloatsep}{6pt}
\begin{table}[!t]
  \centering
  \caption{SGLang throughput (tokens/s; \cref{fig:sglang-throughput}). Green: speedup over AR; bold: best speculative method.}
  \label{tab:sglang-throughput}
\begingroup
\small
\setlength{\tabcolsep}{2.5pt}
\renewcommand{\arraystretch}{0.65}
\newcommand{\sgcell}[2]{#1\,{\scriptsize\textcolor{green!45!black}{#2$\times$}}}
\newcommand{\sgbest}[2]{\textbf{#1}\,{\scriptsize\textcolor{green!45!black}{\textbf{#2\texttimes}}}}
\resizebox{\linewidth}{!}{%
\begin{tabular}{@{}ll*{5}{c}c*{5}{c}@{}}
\toprule
Task & Method & \multicolumn{5}{c}{Qwen3-8B (concurrency)} & & \multicolumn{5}{c}{Qwen3-4B (concurrency)} \\
\cmidrule(lr){3-7}\cmidrule(lr){9-13}
 & & 2 & 4 & 8 & 16 & 32 & & 2 & 4 & 8 & 16 & 32 \\
\midrule
 & Baseline & 381 & 747 & 1405 & 2729 & 4922 &  & 552 & 1072 & 2059 & 3794 & 6768 \\
 & DFlash & \sgcell{1163}{3.05} & \sgcell{2191}{2.93} & \sgcell{3799}{2.70} & \sgcell{5349}{1.96} & \sgcell{6327}{1.29} &  & \sgcell{1615}{2.92} & \sgcell{2991}{2.79} & \sgcell{5250}{2.55} & \sgcell{7873}{2.08} & \sgcell{9534}{1.41} \\
GSM8K & Domino & \sgcell{1193}{3.13} & \sgcell{2229}{2.98} & \sgcell{3884}{2.76} & \sgcell{5576}{2.04} & \sgcell{6675}{1.36} &  & \sgcell{1626}{2.94} & \sgcell{3001}{2.80} & \sgcell{5254}{2.55} & \sgcell{7909}{2.08} & \sgcell{10000}{1.48} \\
 & DSpark & \sgcell{1223}{3.21} & \sgcell{2280}{3.05} & \sgcell{3931}{2.80} & \sgcell{5656}{2.07} & \sgcell{6789}{1.38} &  & \sgcell{1663}{3.01} & \sgcell{3057}{2.85} & \sgcell{5296}{2.57} & \sgcell{7950}{2.10} & \sgcell{10002}{1.48} \\
 & DFlash2 & \sgcell{1327}{3.49} & \sgcell{2473}{3.31} & \sgcell{4234}{3.01} & \sgcell{5899}{2.16} & \sgcell{6975}{1.42} &  & \sgcell{1876}{3.40} & \sgcell{3460}{3.23} & \sgcell{6015}{2.92} & \sgcell{8807}{2.32} & \sgcell{10683}{1.58} \\
 & \textbf{DSpine} & \sgbest{1480}{3.88} & \sgbest{2758}{3.69} & \sgbest{4704}{3.35} & \sgbest{6732}{2.47} & \sgbest{7825}{1.59} &  & \sgbest{2018}{3.65} & \sgbest{3698}{3.45} & \sgbest{6442}{3.13} & \sgbest{9527}{2.51} & \sgbest{11541}{1.71} \\
\midrule
 & Baseline & 380 & 743 & 1394 & 2629 & 4524 &  & 550 & 1061 & 2012 & 3577 & 6107 \\
 & DFlash & \sgcell{1218}{3.21} & \sgcell{2329}{3.13} & \sgcell{4017}{2.88} & \sgcell{5581}{2.12} & \sgcell{6589}{1.46} &  & \sgcell{1673}{3.04} & \sgcell{3144}{2.96} & \sgcell{5605}{2.79} & \sgcell{8192}{2.29} & \sgcell{10115}{1.66} \\
MATH-500 & Domino & \sgcell{1241}{3.27} & \sgcell{2345}{3.16} & \sgcell{4138}{2.97} & \sgcell{5870}{2.23} & \sgcell{6927}{1.53} &  & \sgcell{1682}{3.06} & \sgcell{3156}{2.97} & \sgcell{5584}{2.78} & \sgcell{8474}{2.37} & \sgcell{10708}{1.75} \\
 & DSpark & \sgcell{1260}{3.32} & \sgcell{2382}{3.21} & \sgcell{4136}{2.97} & \sgcell{5914}{2.25} & \sgcell{7032}{1.55} &  & \sgcell{1707}{3.11} & \sgcell{3220}{3.03} & \sgcell{5681}{2.82} & \sgcell{8508}{2.38} & \sgcell{10454}{1.71} \\
 & DFlash2 & \sgcell{1403}{3.69} & \sgcell{2647}{3.56} & \sgcell{4542}{3.26} & \sgcell{6322}{2.40} & \sgcell{7384}{1.63} &  & \sgcell{1991}{3.62} & \sgcell{3691}{3.48} & \sgcell{6452}{3.21} & \sgcell{9638}{2.69} & \sgcell{11453}{1.88} \\
 & \textbf{DSpine} & \sgbest{1522}{4.01} & \sgbest{2875}{3.87} & \sgbest{4968}{3.56} & \sgbest{7030}{2.67} & \sgbest{8200}{1.81} &  & \sgbest{2087}{3.80} & \sgbest{3906}{3.68} & \sgbest{6961}{3.46} & \sgbest{10217}{2.86} & \sgbest{12367}{2.03} \\
\midrule
 & Baseline & 379 & 735 & 1362 & 2541 & 4355 &  & 550 & 1060 & 2027 & 3654 & 5148 \\
 & DFlash & \sgcell{1107}{2.92} & \sgcell{2077}{2.83} & \sgcell{3595}{2.64} & \sgcell{4925}{1.94} & \sgcell{5595}{1.28} &  & \sgcell{1512}{2.75} & \sgcell{2788}{2.63} & \sgcell{4809}{2.37} & \sgcell{7326}{2.00} & \sgcell{8617}{1.67} \\
HumanEval & Domino & \sgcell{1084}{2.86} & \sgcell{2012}{2.74} & \sgcell{3515}{2.58} & \sgcell{5065}{1.99} & \sgcell{5789}{1.33} &  & \sgcell{1494}{2.72} & \sgcell{2784}{2.63} & \sgcell{4714}{2.33} & \sgcell{7250}{1.98} & \sgcell{8757}{1.70} \\
 & DSpark & \sgcell{1126}{2.97} & \sgcell{2081}{2.83} & \sgcell{3618}{2.66} & \sgcell{5088}{2.00} & \sgcell{5861}{1.35} &  & \sgcell{1524}{2.77} & \sgcell{2833}{2.67} & \sgcell{4832}{2.38} & \sgcell{7319}{2.00} & \sgcell{8630}{1.68} \\
 & DFlash2 & \sgcell{1225}{3.23} & \sgcell{2329}{3.17} & \sgcell{3876}{2.85} & \sgcell{5433}{2.14} & \sgcell{6141}{1.41} &  & \sgcell{1755}{3.19} & \sgcell{3347}{3.16} & \sgcell{5488}{2.71} & \sgcell{8334}{2.28} & \sgcell{9515}{1.85} \\
 & \textbf{DSpine} & \sgbest{1354}{3.57} & \sgbest{2541}{3.46} & \sgbest{4325}{3.18} & \sgbest{6158}{2.42} & \sgbest{6963}{1.60} &  & \sgbest{1840}{3.34} & \sgbest{3446}{3.25} & \sgbest{5874}{2.90} & \sgbest{8699}{2.38} & \sgbest{10134}{1.97} \\
\midrule
 & Baseline & 381 & 740 & 1393 & 2682 & 4705 &  & 549 & 1062 & 1996 & 3525 & 5828 \\
 & DFlash & \sgcell{1157}{3.04} & \sgcell{2129}{2.88} & \sgcell{3725}{2.67} & \sgcell{5289}{1.97} & \sgcell{6147}{1.31} &  & \sgcell{1598}{2.91} & \sgcell{2980}{2.81} & \sgcell{5211}{2.61} & \sgcell{7796}{2.21} & \sgcell{9265}{1.59} \\
MBPP & Domino & \sgcell{1139}{2.99} & \sgcell{2117}{2.86} & \sgcell{3677}{2.64} & \sgcell{5338}{1.99} & \sgcell{6302}{1.34} &  & \sgcell{1569}{2.86} & \sgcell{2932}{2.76} & \sgcell{5063}{2.54} & \sgcell{7775}{2.21} & \sgcell{9448}{1.62} \\
 & DSpark & \sgcell{1176}{3.09} & \sgcell{2174}{2.94} & \sgcell{3801}{2.73} & \sgcell{5501}{2.05} & \sgcell{6440}{1.37} &  & \sgcell{1624}{2.96} & \sgcell{3023}{2.85} & \sgcell{5274}{2.64} & \sgcell{7803}{2.21} & \sgcell{9515}{1.63} \\
 & DFlash2 & \sgcell{1275}{3.35} & \sgcell{2333}{3.15} & \sgcell{4042}{2.90} & \sgcell{5667}{2.11} & \sgcell{6646}{1.41} &  & \sgcell{1830}{3.33} & \sgcell{3376}{3.18} & \sgcell{6064}{3.04} & \sgcell{8771}{2.49} & \sgcell{10287}{1.76} \\
 & \textbf{DSpine} & \sgbest{1399}{3.68} & \sgbest{2586}{3.50} & \sgbest{4465}{3.21} & \sgbest{6411}{2.39} & \sgbest{7321}{1.56} &  & \sgbest{1948}{3.55} & \sgbest{3673}{3.46} & \sgbest{6346}{3.18} & \sgbest{9369}{2.66} & \sgbest{11072}{1.90} \\
\bottomrule
\end{tabular}%
}
\endgroup
\end{table}

\paragraph{Evaluation.}
Each benchmark uses its complete test set (1,319 GSM8K, 500 MATH-500, 164 HumanEval, 500 MBPP, 1,055 LCB, 80 MT-Bench, and 500 Arena-Hard prompts; MT-Bench runs both turns). All methods receive identical prompts without a system message and generate at most 2,048 new tokens. Acceptance length in \cref{tab:main-acceptance} is measured in SGLang with eight single-GPU instances, each serving up to 16 concurrent requests; at temperature one, tokens are sampled from the full target distribution. Every method proposes 15 draft tokens per round. $\tau$ is averaged over the responses of each benchmark, and Avg.\ is the unweighted mean over benchmarks.

\paragraph{Lossless verification.}
If the first rejection occurs at position $j$, we keep $\hat{x}_{1:j-1}$ and sample a replacement from
\begin{equation}
p_j^{\mathrm{corr}}(v)=
\frac{[p_j(v)-q_{\mathrm{draft},j}(v)]_+}
{\sum_{u\in\mathcal V}[p_j(u)-q_{\mathrm{draft},j}(u)]_+},
\qquad [a]_+=\max(a,0),
\label{eq:spec-correction}
\end{equation}
where $\mathcal V$ is the vocabulary; if all $M$ candidates are accepted, one additional token is sampled from the target. Together with \cref{eq:spec-accept}, this preserves the target distribution.

\paragraph{Decoding at temperature one.}
All methods use SGLang's lossless verification. DSpark, DFlash2, and DSpine sample their proposals, the latter two from the softmax of their selector or transition-cache scores over the top-16 candidates of each position, and apply rejection sampling with these proposal probabilities; DFlash and Domino propose greedily and accept a candidate when it matches the token sampled from the target.

\paragraph{Serving.}
Each instance uses one GPU (tensor parallelism one, BF16, FlashAttention-3 \citep{shah2024flashattention3}, CUDA graphs) at temperature zero with at most 2,048 output tokens on the complete GSM8K, MATH-500, HumanEval, and MBPP test sets, with identical prompts in the same order. Throughput is the total number of output tokens divided by the wall-clock time of the whole workload, including prefill and scheduling. All methods use the same static budget of 15 draft tokens per round.

\paragraph{Supervision targets.}
The variants in \cref{tab:supervision-ablation} use the same training data, disable injection, and are trained for three epochs (1,827 steps); token CE decodes every layer through the shared final RMSNorm and the frozen target LM head. All variants are evaluated in SGLang at temperature zero.

\end{document}